\documentclass[runningheads]{llncs}

\usepackage[T1]{fontenc}
\usepackage[utf8]{inputenc}
\usepackage{graphicx}
\usepackage{booktabs}
\usepackage{tabularx}
\usepackage{array}
\usepackage{multirow}
\usepackage{amsmath}    
\usepackage{amssymb}
\usepackage{microtype}
\usepackage{xcolor}
\usepackage{tikz}
\usepackage{url}
\usepackage{float}
\usepackage{comment}
\usepackage[section]{placeins}
\usetikzlibrary{arrows.meta,positioning,fit,calc}
\graphicspath{{figures/}{./}}

\newcolumntype{Y}{>{\raggedright\arraybackslash}X}
\newcolumntype{C}{>{\centering\arraybackslash}X}
\newcommand{\na}{\textemdash}
\newcommand{\iou}{\operatorname{IoU}}
\newcommand{\modelname}{G-Unified}

\title{Learning to Ground Before Reading: Unified PCB Engineering Drawing Parsing with Compact Vision-Language Models}
\titlerunning{Unified Detector-Free Grounding of PCB Drawings}

\author{Jinghao Liu\inst{2}\thanks{These authors contributed equally to this work.} \and Xingrun Liu\inst{2,*} \and
Gengchen Sun\inst{1,2,3,4} \and Han Xiao\inst{4} \and
Xingyu Chen\inst{4} \and Yuhui Deng\inst{1,2}}
\authorrunning{Authors Suppressed Due to Excessive Length}
\institute{
Guangdong Provincial Key Laboratory of Interdisciplinary Research and Application for Data Science, Zhuhai 519087, China \and
Beijing Normal--Hong Kong Baptist University, Zhuhai 519087, China \and
Faculty of Science, Hong Kong Baptist University, Hong Kong SAR 999077, China \and
Hong Kong aiKnow Limited, Hong Kong SAR, China \\
\email{\{sungengchen, ivandeng\}@bnbu.edu.cn}}

\begin{document}

\maketitle
\begin{abstract}
PCB engineering drawings mix sparse graphics, dense tables, and text whose meaning depends on page position. Localizing the regions and sending crops to specialized recognizers are determined as the methods for most parsers, so missed regions cannot be recovered downstream. We train a compact VLM to read the full page and get a sequence of region classes, normalized boxes, and text or HTML content. Bounding boxes are converted to coordinate tokens for supervision. Inference uses no detector or crop parser. The joint target is difficult to optimize because class and box tokens are sparse relative to the much longer content sequences. Our localization-first curriculum learns the class--box format before adding content targets with content-aware resampling. On the fixed validation split of the Engineering Drawing Dataset (ED dataset), Localization-First improves strict localization F1 by 0.0955 over joint training (paired image-bootstrap 95\% interval: $[0.0350,0.1572]$). G-Unified has the lowest NED, highest cell F1, and only nonzero exact-match score. It provides a detector-free baseline for full-page PCB drawing parsing.

\end{abstract}

\section{Introduction}

PCB engineering drawings are usually highly informative, including graphical views, fabrication notes, layer stackups, drill schedules, and heterogeneous tables on a single high-resolution page. Experienced people can easily locate the wanted information, however, it is hard for the vision model, as this is more than the recognition problem: content must be interpreted together with its spatial context and semantic role. Beyond identifying where content occurs, explicit localization recovers the page layout required for downstream tracing, because relationships among drawing elements often have to be inferred from their relative positions and spatial arrangement rather than from content alone.A complete parser therefore needs to identify relevant regions, determine what each region represents, recover its textual or tabular content, and retain its location on the page. 

Most existing document and engineering-drawing systems address these requirements with a cascaded pipeline. A detector or layout analyzer first localizes regions, after which individual crops are routed to Optical Character Recognition(OCR) engines or Vision-Language Models for recognition. Systems such as eDOCr2 combine geometry-driven segmentation with specialized recognition modules \cite{Toro2025}, while Hybrid-VL uses a trained YOLO11m-OBB detector before parsing detected crops with a VLM \cite{Khan2025HybridVL,Kim2022Donut,Xiao2024Florence2}. This decomposition creates a hard dependency between localization and recognition: a missed or incorrectly cropped region cannot be recovered by the downstream parser. It also introduces detector-specific training, crop-routing rules, and multiple inference stages. 

We investigate whether this cascade can be replaced by a single compact VLM operating directly on the full page. Our Unified Grounding Training framework (\modelname{}) formulates PCB parsing as structured sequence generation, where each predicted region is represented by its semantic class, bounding box, and textual or HTML content. The model therefore performs classification, localization, and parsing jointly in one autoregressive call, without detector inference, crop routing, or a second recognizer. 

This unified formulation, however, introduces an optimization imbalance. Class and coordinate fields are short and highly structured, whereas text and table contents are substantially longer and more variable, causing ordinary joint training to be dominated by content tokens. We address this issue with \textit{Localization-First} training, which first establishes region classes and locations before introducing the complete \texttt{\{class, box, content\}} target. In general, our contributions can be illustrated as follows: 

\begin{enumerate} 
    \item \textbf{Unified full-page grounding and parsing.} We formulate PCB engineering-drawing parsing as a single \texttt{\{class, box, content\}} generation task and train a compact VLM to jointly classify, localize, and parse regions directly from the full page, eliminating detector inference and crop-based recognition. 
    \item \textbf{Localization-First training.} We introduce a two-phase training strategy that prioritizes sparse localization supervision before the longer content targets, improving the optimization of unified structured generation. 
\end{enumerate}

\section{Related Work}

Document understanding has evolved from explicit layout analysis toward multimodal and generative page modeling. DocLayNet provides large-scale layout annotations \cite{Pfitzmann2022}, while LayoutParser offers reusable components for region detection and segmentation \cite{Shen2021}. LayoutLM and LayoutLMv3 combine textual, spatial, and visual features but typically rely on OCR tokens or task-specific prediction heads \cite{Xu2020,Huang2022LayoutLMv3}. OCR-free models such as Donut, Pix2Struct, and Florence-2 instead generate structured outputs directly from page pixels \cite{Kim2022Donut,Lee2023Pix2Struct,Xiao2024Florence2}, while table-oriented work including PubTabNet, PubTables-1M, and UniTabNet studies the recovery of textual and structural information from document images \cite{Zhong2020PubTabNet,Smock2022PubTables,Zhang2024UniTabNet}. Generative VLMs further extend this paradigm to localization: Qwen-VL introduced referential grounding alongside text understanding \cite{Bai2023QwenVL}, later Qwen variants improved variable-resolution perception and document understanding \cite{Wang2024Qwen2VL,Bai2025Qwen25VL,Bai2025Qwen3VL}, and Kosmos-2 links generated language to image regions \cite{Peng2024Kosmos2}. Recent page parsers either serialize layout and content directly, as in dots.ocr \cite{Li2025Dots}, or adopt coarse-to-fine processing for high-resolution pages, as in MinerU2.5 and Intelligent Document Parsing \cite{Niu2025MinerU,Xing2025IDP}.

Engineering drawings remain particularly challenging because they combine sparse line graphics, rotated annotations, dense tables, and domain-specific layouts. Prior work has explored VLM-based information extraction from engineering drawings \cite{Khan2024Florence,Picard2024Concept}, symbol recognition in blueprints \cite{Shteriyanov2025Blueprint}, and multimodal reasoning over architectural and engineering documents \cite{AECVBench2026}. More specialized systems commonly use cascaded pipelines: eDOCr2 applies geometric heuristics before dedicated recognition modules \cite{Toro2025}, while Hybrid-VL first localizes regions with a rotation-aware YOLO detector and then parses the resulting crops with a VLM \cite{Khan2025HybridVL}. In contrast, our work studies whether region classification, localization, and content extraction can be unified as full-page structured generation with a compact VLM. We additionally use parameter-efficient adaptation through LoRA/QLoRA \cite{Hu2022,Dettmers2023} and a curriculum-inspired training strategy \cite{Bengio2009Curriculum} that introduces localization supervision before the complete structured target.

\section{Methodology}

\begin{figure}[htp]
    \centering
    \includegraphics[width=\linewidth]{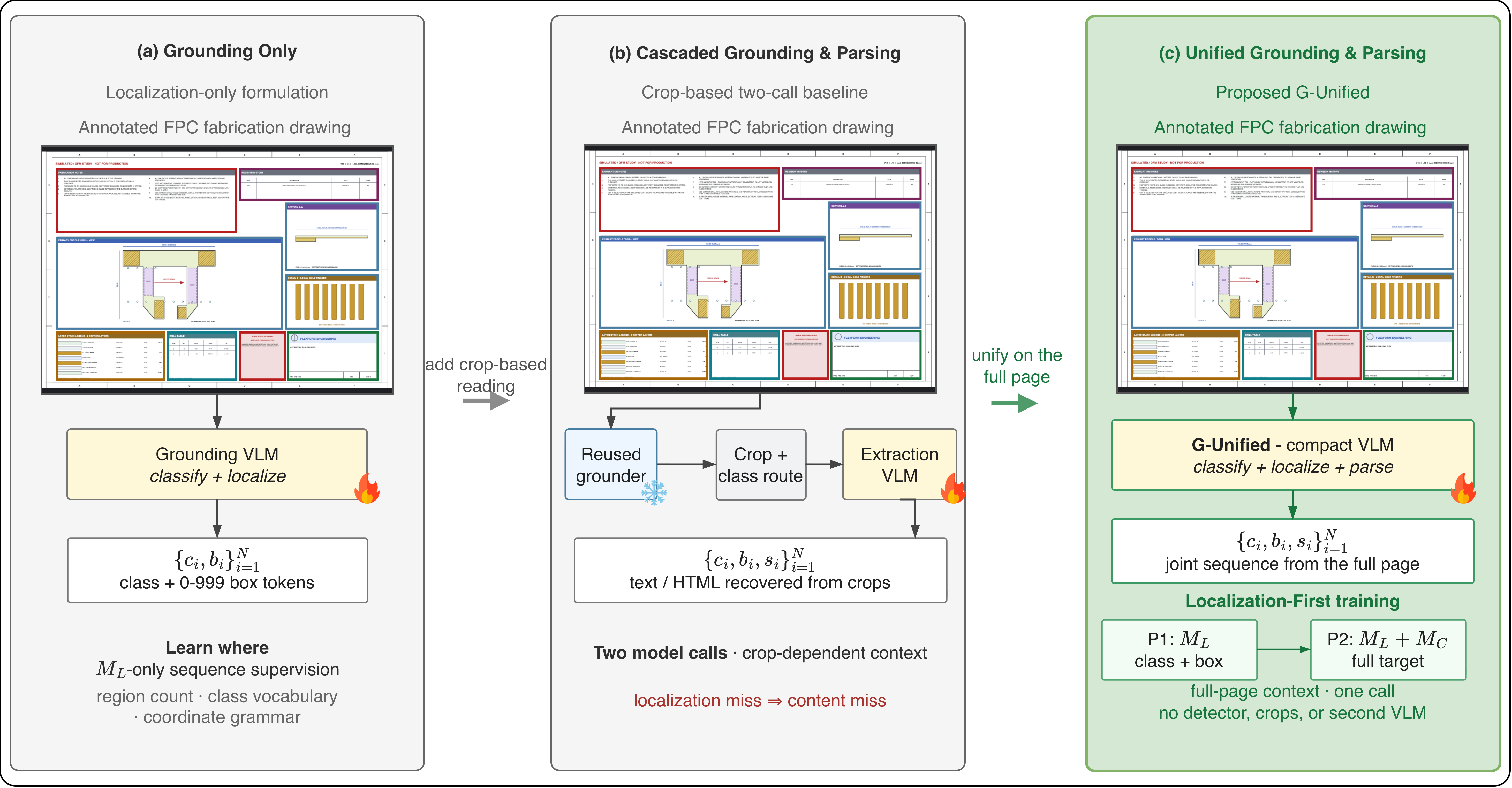}
    \caption{Three parsing modes considered in this study. (a) Grounding Only predicts region classes and locations without content. (b) Cascaded Grounding \& Parsing first localizes and crops each region, then sends the crop to a separate VLM. (c) Unified Grounding \& Parsing predicts classes, boxes, and content directly from the full page in one VLM call. Green highlights the unified mode used in the primary experiments.}
    \label{fig:modes}
\end{figure}

Examining how localization and content extraction should be coupled, Figure~\ref{fig:modes} compares three parsing modes under the same region representation. The primary experiments on the ED dataset use the unified mode implemented by \modelname{}; the other two serve as localization-only and cascaded references.

\subsection{Unified Output Representation}

Let a rasterized drawing be $I$ and let the six-class ontology be
\[
\mathcal{C}=\{\text{text},\text{figure},\text{title block},\text{drill},\text{stackup},\text{other table}\}.
\]
The target page representation is
\begin{equation}
\mathcal{U}(I)=\{(c_i,b_i,s_i)\}_{i=1}^{N},
\qquad c_i\in\mathcal{C},\quad b_i\in\{0,\ldots,999\}^{4},
\label{eq:task}
\end{equation}
where $b_i=(x_{1i},y_{1i},x_{2i},y_{2i})$ is an axis-aligned box normalized to the page and $s_i$ is plain text, an HTML table, or an empty string for graphical figures. Fixed-grid coordinates make geometry independent of raster resolution and permit deterministic parsing.

Each object is serialized as a class header, four box integers, and a content field. Page targets concatenate object records in reading-compatible source order.

\subsection{Detector-Free One-Call Parsing}

For a two-stage system, successful content recovery requires both a matched localization and a correct crop parse. If $A_i$ denotes localization success and $B_i$ denotes correct parsing conditional on that crop, then
\begin{equation}
P(A_i\cap B_i)=P(A_i)P(B_i\mid A_i).
\label{eq:cascade}
\end{equation}
A localization miss therefore forces a content miss in the two-stage pipeline, even when the crop recognizer itself is accurate. Unified generation still couples recognition and localization, but all output tokens attend to the full-page representation and preceding page context instead of a detector-selected crop.

\subsection{Joint Sequence Objective}

Given visual tokens $v(I)$ and serialized target tokens $y_{1:T}$, the model minimizes teacher-forced negative log likelihood
\begin{equation}
\mathcal{L}_{\mathrm{joint}}(I,y)
=-\sum_{t=1}^{T}m_t\log p_{\theta}(y_t\mid v(I),y_{<t}),
\label{eq:jointloss}
\end{equation}
where $m_t$ masks prompt and padding positions. The shared likelihood trains localization, classification, and parsing together. Because content fields contain far more tokens than boxes and labels, they contribute most loss terms. Long HTML targets can also postpone later regions in the sequence.

\subsection{Localization-First Training}

We split each target mask into localization/class tokens $M_L$ and content tokens $M_C$. Phase 1 retains only class and box fields:
\begin{equation}
\mathcal{L}_{1}=-\sum_{t}M_{L,t}\log p_{\theta}(y_t\mid v(I),y_{<t}).
\label{eq:phase1}
\end{equation}
Phase 2 restores the complete output grammar:
\begin{equation}
\mathcal{L}_{2}=-\sum_{t}(M_{L,t}+M_{C,t})
\log p_{\theta}(y_t\mid v(I),y_{<t}).
\label{eq:phase2}
\end{equation}
Phase 1 trains region counting, class vocabulary, coordinate syntax, and stopping behavior. Phase 2 adds the content fields while retaining class and box supervision. The schedule requires no manually tuned coefficient between localization and content losses.

\section{Experiments}

\subsection{Dataset and Fixed Split}

The ED dataset contains 267 drawings and 2,209 annotated regions. Its frozen split has 241 training drawings with 1,937 regions and 26 validation drawings with 272 regions. Among all regions, 522 have aligned content supervision: 444 for training and 78 for validation. The 78 nonempty validation targets occur on 20 of the 26 drawings. The validation content consists of 56 text notes, 14 title blocks, three drill tables, three other tables, and two stackup tables. The same image identifiers, boxes, six-class mapping, and content references are used for every cross-system result. We report sampled-record counts separately because resampling changes optimization frequency but not the number of distinct drawings.

\subsection{Training Variants and Implementation Details}

\paragraph{Variants.}

\begin{table}[h]
\caption{Unified-training variants. Counts are records sampled per epoch, not distinct drawings. All rows use the same validation split of the ED dataset.}
\label{tab:variants}
\centering
\small
\begin{tabularx}{\textwidth}{lccY}
\toprule
Variant & Backbone & Records/epoch & Training intervention \\
\midrule
2B unified & 2B & 241 & Original joint target \\
4B baseline & 4B & 241 & Original joint target \\
4B no-thinking & 4B & 241 & Direct structured answer only \\
4B stable & 4B & 802 & Content-heavy continued adaptation \\
4B Local-First & 4B & 241 / 615 & Box-only warm-up, then full targets \\
4B hard-class & 4B & 961 & Hard-label sampling and dirty-content removal \\
\bottomrule
\end{tabularx}
\end{table}

The 2B and 4B baseline rows differ only in model capacity, as shown in Table 1, which summarizes all unified-model training variants.The continued runs start from different adapters and use different target mixtures, so their absolute losses are reported only as optimization traces and are not compared directly across variants.

\paragraph{Localization-First schedule.}
The ED dataset contains 241 page conversations: 187 include at least one content target and 54 contain boxes only. Phase 1 uses 241 box-only targets for two epochs at learning rate $5\times10^{-5}$. Phase 2 starts from that adapter, restores the full targets, and samples each content-bearing page twice in addition to its regular occurrence. The resulting per-epoch sample count is
\begin{equation}
241+2(187)=615
\end{equation}
for each of three epochs at learning rate $3\times10^{-5}$. Repetition changes the sampling distribution but does not add new drawings.

\begin{figure}[!t]
  \centering
  \includegraphics[width=\linewidth]{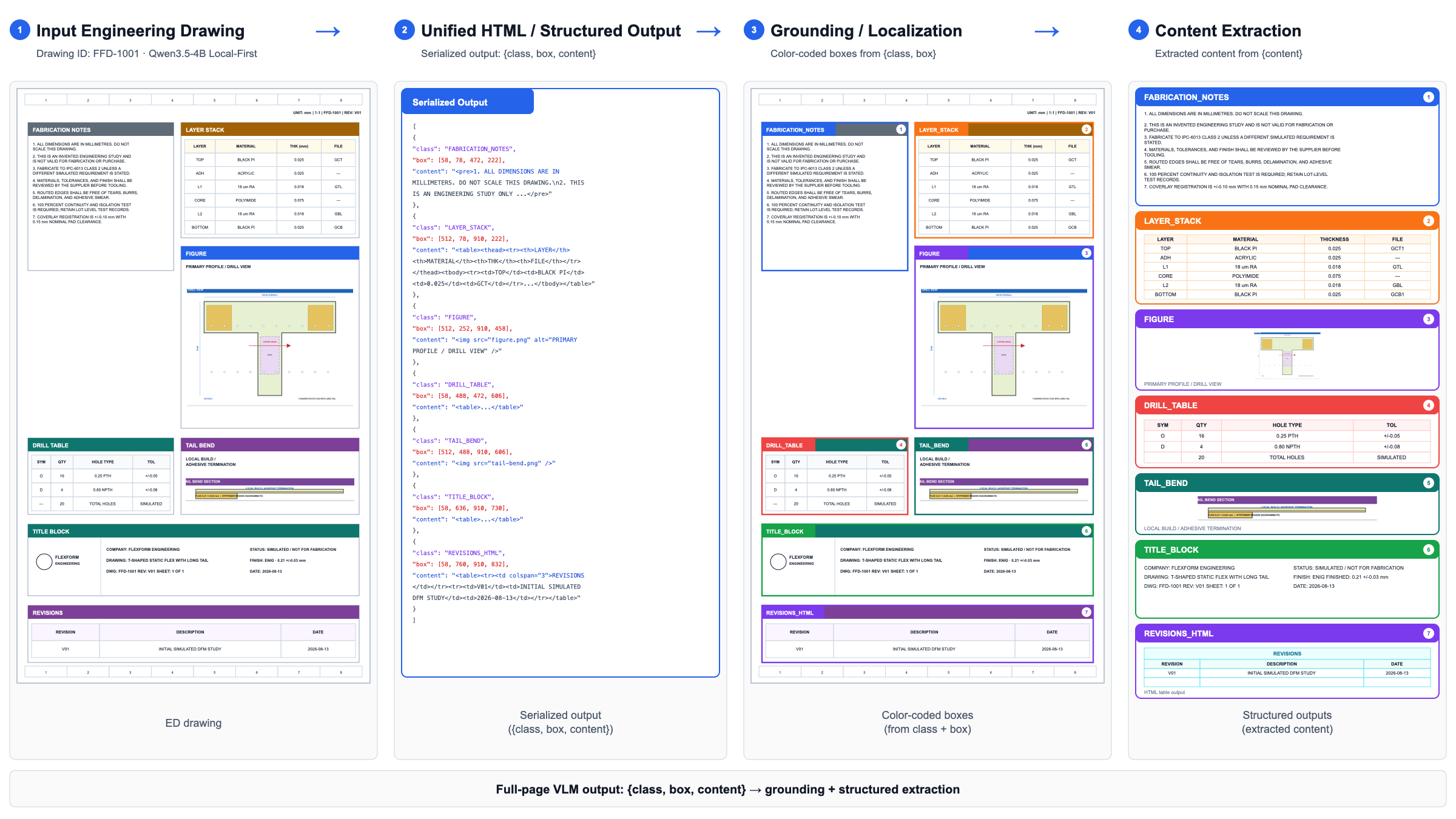}
  \caption{End-to-end Local-First parsing with full-page localization and structured content extraction.}
  \label{fig:end-to-end-example}
\end{figure}

\paragraph{Implementation.}
The principal model is Qwen3.5-4B; the 2B counterpart uses the same grammar and preprocessing. The 4B base is quantized to 4-bit precision and adapted with LoRA rank 32, scaling 64, and dropout 0.05. LoRA modules cover the query, key, value, output, gate, up, and down projections. The maximum sequence length is 8,192 tokens, the image budget is 1,048,576 pixels, the micro-batch size is one, and gradient accumulation is 16. Decoding is deterministic, with thinking disabled and at most 6,144 new tokens. This configuration fits page-level training within the available memory while retaining sufficient resolution for dense PCB annotations. Figure~\ref{fig:end-to-end-example} shows the corresponding full-page localization and structured outputs. We cap each object’s content at 1,200 characters and the complete assistant target at 6,000 characters, preserving a closing </table> tag whenever possible. Due to data privacy, the PCB model displayed here is generated by virtual engineering drawing.

\subsection{Reproduced Cascade Baselines}

The eDOCr2 reproduction calibrates frame, GDT, and binarization thresholds on the 241 training drawings, then freezes them for validation. Geometry and contour rules propose regions; text crops are read with Tesseract, while table crops are sent to Qwen2-VL-7B in 4-bit mode. The selected thresholds are 0.65, 0.015, and 127, respectively. No validation image is used for calibration. After restoring the CUDA 12.8 NVRTC dependency, we rerun only content extraction on the frozen validation detections; localization outputs are unchanged.

The Hybrid-VL reproduction fine-tunes YOLO11m-OBB for 10 epochs at $1024\times1024$ resolution, batch size 8, automatic mixed precision, and seed 42. For the common localization protocol, each predicted oriented box is converted to its enclosing axis-aligned box. Its second stage fully fine-tunes Florence-2-base-ft on 444 human-annotated crops from the ED dataset for 10 epochs with AdamW, a cosine schedule, learning rate $10^{-6}$, batch size 1, and gradient accumulation 8. The original Hybrid-VL study uses longer 400/30-epoch detector/recognizer schedules \cite{Khan2025HybridVL}; our results therefore characterize the supplied 10/10-epoch reproduction, not the upper limit of the architecture.

\subsection{Evaluation}\label{sec:evaluation}

\paragraph{Localization.}
Predictions are matched one-to-one with references within the same canonical class at $\iou\geq0.5$. Let $TP$, $FP$, and $FN$ be the resulting micro counts. We report
\begin{equation}
P=\frac{TP}{TP+FP},\quad
R=\frac{TP}{TP+FN},\quad
F_1=\frac{2PR}{P+R},
\label{eq:metrics}
\end{equation}
and mean IoU over matched pairs. Every emitted item is counted. A record with a non-canonical raw label remains a false positive; it is not dropped or remapped after generation. For uncertainty, we resample the 26 validation drawings with replacement and recompute paired micro-F1 differences for $B=5{,}000$ replicates. The reported interval is the percentile 95\% interval. Pairing preserves the fact that every model is evaluated on the same pages and is preferable here to treating boxes as independent observations.

\paragraph{Content extraction.}
Content is associated only through the strict class-aware localization match. Normalized edit distance is
\begin{equation}
\operatorname{NED}(r,h)=
\frac{d_{\mathrm{lev}}(r,h)}{\max(1,|r|,|h|)},
\label{eq:ned}
\end{equation}
where $r$ and $h$ are reference and hypothesis strings. For HTML tables, cell precision, recall, and F1 are computed after structured parsing; exact match tests normalized text identity. An unmatched content reference receives NED 1 and zero cell or exact-match credit. Malformed outputs that the configured pipeline cannot parse are treated as empty; we do not repair raw generations. Assigning zero credit to unmatched or unusable content makes the score end to end, unlike recognition measured only on oracle crops.

\section{Results}

\subsection{Controlled Unified-Model Comparison}

Table~\ref{tab:unified-results} reports held-out results for the unified models. Increasing the backbone from 2B to 4B raises F1 from 0.1948 to 0.3942. The difference is consistent with higher capacity helping on long joint targets, although only one trained run is available for each size. Removing reasoning traces improves recall and F1 but slightly worsens NED. Stable continued adaptation improves localization, while its heavier content sampling does not improve content similarity. Local-First reaches precision 0.6485, F1 0.4897, matched IoU 0.8907, and NED 0.8143. Compared with the 4B baseline, F1 increases by 0.0955 (24.2\% relative) and NED decreases by 0.0682. Hard-class has the highest F1 (0.4989) and recall (0.4081), but its NED is higher at 0.8466. No variant leads every localization and content metric.

\begin{table}[htp]
\caption{Strict one-pass results on the same 26 validation drawings from the ED dataset (272 boxes; 78 content-bearing references). NED is lower-is-better; all other metrics are higher-is-better. Missing content receives NED 1 and zero task credit.}
\label{tab:unified-results}
\centering
\scriptsize
\setlength{\tabcolsep}{3.4pt}
\begin{tabular}{lrrrrrrrr}
\toprule
Model & Pred. & Prec. & Rec. & $F_1$ & mIoU & NED$\downarrow$ & Cell $F_1$ & EM \\
\midrule
2B unified & 231 & .2121 & .1801 & .1948 & .6982 & \na & \na & \na \\
4B baseline & 144 & .5694 & .3015 & .3942 & .8083 & .8825 & .1464 & .0000 \\
4B no-thinking & 191 & .5550 & .3897 & .4579 & .8293 & .8629 & \textbf{.2230} & .0000 \\
4B stable & 188 & .5798 & .4007 & .4739 & .8751 & .8347 & .1497 & .0145 \\
4B Local-First & 165 & \textbf{.6485} & .3934 & .4897 & .8907 & \textbf{.8143} & .1516 & \textbf{.0299} \\
4B hard-class & 173 & .6416 & \textbf{.4081} & \textbf{.4989} & \textbf{.8982} & .8466 & .1945 & .0294 \\
\bottomrule
\end{tabular}
\end{table}

\begin{figure}[h]
  \centering
  \includegraphics[width=\textwidth]{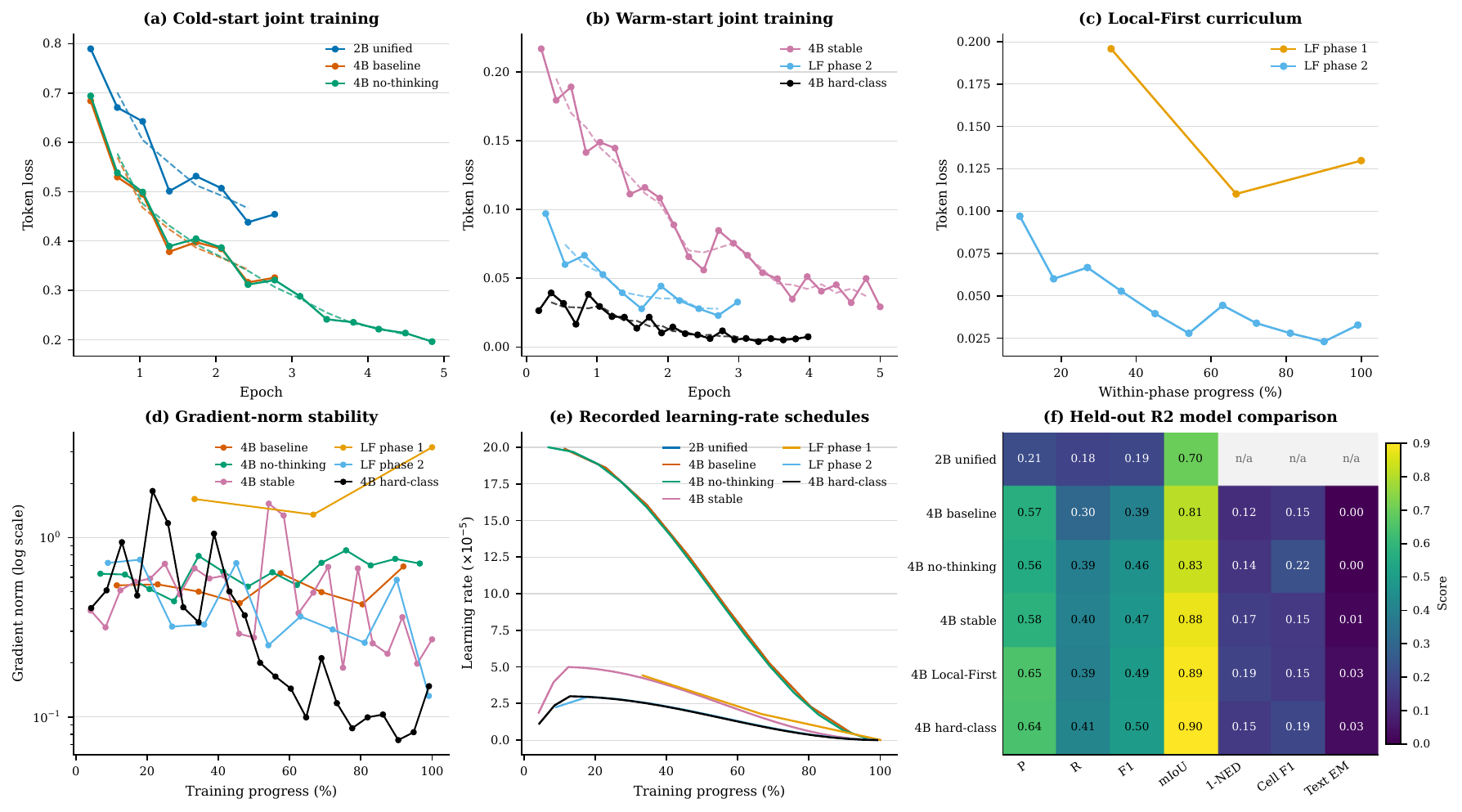}
  \caption{Unified-model training and held-out comparison. (a) Cold-start joint losses for 2B and 4B models. (b) Warm-start and continued-run losses. (c) The two Local-First phases. (d) Gradient norms and (e) learning-rate schedules. (f) Held-out metrics on the ED dataset, with sequence similarity shown as $1-\mathrm{NED}$. Panels (a)--(e) show optimization traces; panel (f) reports held-out scores.}
  \label{fig:training}
\end{figure}

Training loss decreases in every run. Cold-start 4B loss falls from 0.6842 to 0.3259, no-thinking from 0.6941 to 0.1963, and the 2B run from 0.7897 to 0.4543. Local-First phase 1 ends at 0.1298 and phase 2 at 0.0327. Learning rates decay as scheduled, and no run shows sustained divergence. Absolute losses are not comparable across warm- and cold-start runs because their initial adapters and token mixtures differ. Held-out metrics in panel (f) are used for model comparison.

\subsection{Paired Uncertainty and Class Behavior}

Paired bootstrap results are reported in Table~\ref{tab:bootstrap}. For Local-First minus the 4B baseline, the 95\% interval excludes zero and 99.86\% of paired replicates are positive. The intervals for Local-First minus stable and hard-class minus Local-First include zero, so the 26-page split does not distinguish either pair.

\begin{table}[h]
\caption{Paired image bootstrap for strict micro-F1 on the ED dataset ($B=5{,}000$). The last column is the fraction of replicates with positive difference, not a parametric $p$-value.}
\label{tab:bootstrap}
\centering
\small
\begin{tabular}{lrrr}
\toprule
Paired contrast & $\Delta F_1$ & 95\% interval & $\Pr(\Delta>0)$ \\
\midrule
Local-First $-$ baseline & .0955 & $[.0350,.1572]$ & .9986 \\
Local-First $-$ stable & .0158 & $[-.0598,.0862]$ & .6346 \\
Hard-class $-$ Local-First & .0092 & $[-.0428,.0551]$ & .6430 \\
\bottomrule
\end{tabular}
\end{table}

\begin{figure}[h]
  \centering
  \includegraphics[width=\textwidth]{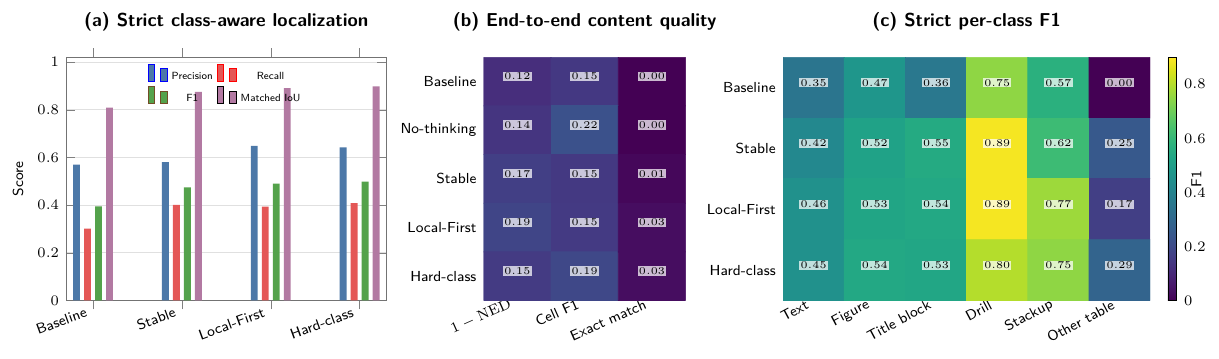}
  \caption{Strict diagnostics on the ED dataset. (a) Localization precision, recall, F1, and matched IoU. (b) Full-denominator content metrics, with $1-\mathrm{NED}$ used so every cell is higher-is-better. (c) Per-class F1. Paired bootstrap intervals are reported in Table~\ref{tab:bootstrap}.}
  \label{fig:summary}
\end{figure}

Per-class F1 follows the same general pattern. Relative to the baseline, Local-First improves text from 0.3488 to 0.4551, figure from 0.4741 to 0.5306, title block from 0.3611 to 0.5405, and stackup from 0.5714 to 0.7692; drill remains 0.8889. Hard-class improves other-table F1 from 0 to 0.2857 but lowers drill and stackup scores relative to Local-First. Several classes contain few references, so their percentages should be interpreted cautiously.

\subsection{Content Recall and Output Grammar}

Full-denominator content scores remain below scores computed only on matched regions. Local-First matches 27 of the 78 content-bearing references. On those matches, it obtains NED 0.4634, cell F1 0.3169, and exact match 0.1250; 51 references are unmatched and four matched outputs are empty. Hard-class matches 25 references, reaches NED 0.5215 and cell F1 0.4278 on the matches, and leaves 53 references unmatched, including eight empty matched outputs. Most end-to-end content loss comes from unmatched regions, followed by errors in the recovered content. Strict parsing also reveals invalid output labels. Local-First emits 15 non-canonical labels among 165 predictions (9.1\%), and hard-class emits 10 among 173 (5.8\%); all are counted as false positives. 

\subsection{Same-Split Cross-System Localization}

Table~\ref{tab:cross-loc} places the three runnable system families on the same validation split of the ED dataset. Hybrid-VL has the highest precision, recall, and F1. Local-First is 0.1807 F1 lower, but it has the highest mean IoU among matched boxes and requires neither detector inference nor crop routing. eDOCr2 detects large drawing views reasonably often but misses most text and semantic table classes.

\begin{table}[h]
\caption{Strict localization on the same 26 validation drawings from the ED dataset (272 boxes). Every emitted record is counted; higher is better.}
\label{tab:cross-loc}
\centering
\small
\begin{tabular}{lrrrrr}
\toprule
System & Pred. & Precision & Recall & $F_1$ & mIoU \\
\midrule
eDOCr2 reproduction \cite{Toro2025} & 214 & .2383 & .1875 & .2099 & .8280 \\
\modelname{} Local-First & 165 & .6485 & .3934 & .4897 & \textbf{.8907} \\
Hybrid-VL reproduction \cite{Khan2025HybridVL} & 268 & \textbf{.6754} & \textbf{.6654} & \textbf{.6704} & .8215 \\
\bottomrule
\end{tabular}
\end{table}

Both Hybrid-VL stages optimize stably under the 10/10-epoch reproduction, although the short schedule does not establish convergence to the original 400/30-epoch setting.

\subsection{Cross-System Content Extraction}

Table~\ref{tab:cross-content} extends the comparison to content. The ``localized'' column counts references with a same-class localization match, while ``usable'' additionally requires a nonempty output accepted by the configured pipeline. Content scores follow the full-denominator protocol in Section~\ref{sec:evaluation}. The Hybrid-VL entry in Table~\ref{tab:cross-content} reports the unmodified saved pipeline. Repair-only and prefix-regenerate are post-hoc decoder diagnostics; neither changes the detector or Florence weights, and neither is substituted into the main comparison.

\begin{table}[h]
\caption{Strict end-to-end content extraction on the same 78 content references from 20 content-bearing drawings in the ED dataset. NED is lower-is-better; cell F1 and exact match (EM) are higher-is-better.}
\label{tab:cross-content}
\centering
\scriptsize
\setlength{\tabcolsep}{3.8pt}
\begin{tabular}{lcccrrr}
\toprule
System & Content stage & Localized & Usable & NED$\downarrow$ & Cell $F_1$ & EM \\
\midrule
eDOCr2 & Tesseract + Qwen2-VL-7B & 13/78 & 9/78 & .9856 & .0602 & .0000 \\
Hybrid-VL & Florence-2-base-ft & \textbf{48/78} & 0/78 & 1.0000 & .0000 & .0000 \\
\modelname{} Local-First & page-level Qwen3.5-4B & 27/78 & \textbf{23/78} & \textbf{.8143} & \textbf{.1516} & \textbf{.0299} \\
\bottomrule
\end{tabular}
\end{table}

The repaired eDOCr2 run writes content records for all 74 content-like detections, yet only 13 of 78 references receive a same-class match at $\iou\geq0.5$ and nine produce usable content. Its Qwen2-VL table stage completes for 27 predicted regions without a table-stage exception; 47 predicted records still carry extraction errors. The 65 unmatched references alone contribute $65/78=0.8333$ to full-denominator NED. The final NED of 0.9856, cell F1 of 0.0602, and zero EM show that localization is the dominant error source and the matched content also remains inaccurate.
Hybrid-VL executes both stages: YOLO localization followed by Florence-2 content extraction. The detector provides same-class matches for 48 of 78 content references, but the unmodified saved Florence-2 outputs have JSON format rate 0. Under strict end-to-end scoring, none is usable, giving NED 1.0000, cell F1 0, and EM 0.

To separate malformed serialization from recognition, we run two post-hoc diagnostics on 78 Florence generations produced from human-annotated crops in the ED dataset. Repair-only strips code fences, parses valid or embedded JSON, recovers text after observed category/content markers, rejects near-empty strings, and rebuilds the two-field schema. Prefix-regenerate reloads the same 10-epoch checkpoint, seeds the decoder with a class-conditioned category/content prefix, greedily generates only the content continuation, and applies the same repair fallback. We define format rate as $R_{\mathrm{fmt}}=N^{-1}\sum_i \mathbf{1}[\operatorname{parse}(y_i)\neq\emptyset]$.
\begin{table}[h]
\caption{Florence decoder diagnostics on 78 human-annotated crops from the ED dataset.}
\label{tab:florence-format}
\centering
\small
\setlength{\tabcolsep}{5pt}
\begin{tabular}{lrrrr}
\toprule
Decode & Format rate & Usable & Mean NED$\downarrow$ & EM \\
\midrule
Original saved output & .000 & 0/78 & 1.000 & .000 \\
Repair-only & 1.000 & 78/78 & \textbf{.515} & .000 \\
Prefix-regenerate + repair & 1.000 & 78/78 & .553 & .000 \\
\bottomrule
\end{tabular}
\end{table}
Both diagnostics achieve a format rate of 1.0, but content accuracy remains low. Repair-only gives a mean NED of 0.515 and a structured-table cell F1 of 0.1646, while prefix-regenerate gives a mean NED of 0.553. EM remains zero for both. G-Unified therefore remains strongest on content among the unmodified runs.

\begin{figure}[htp]
  \centering
  \includegraphics[width=0.8\textwidth]{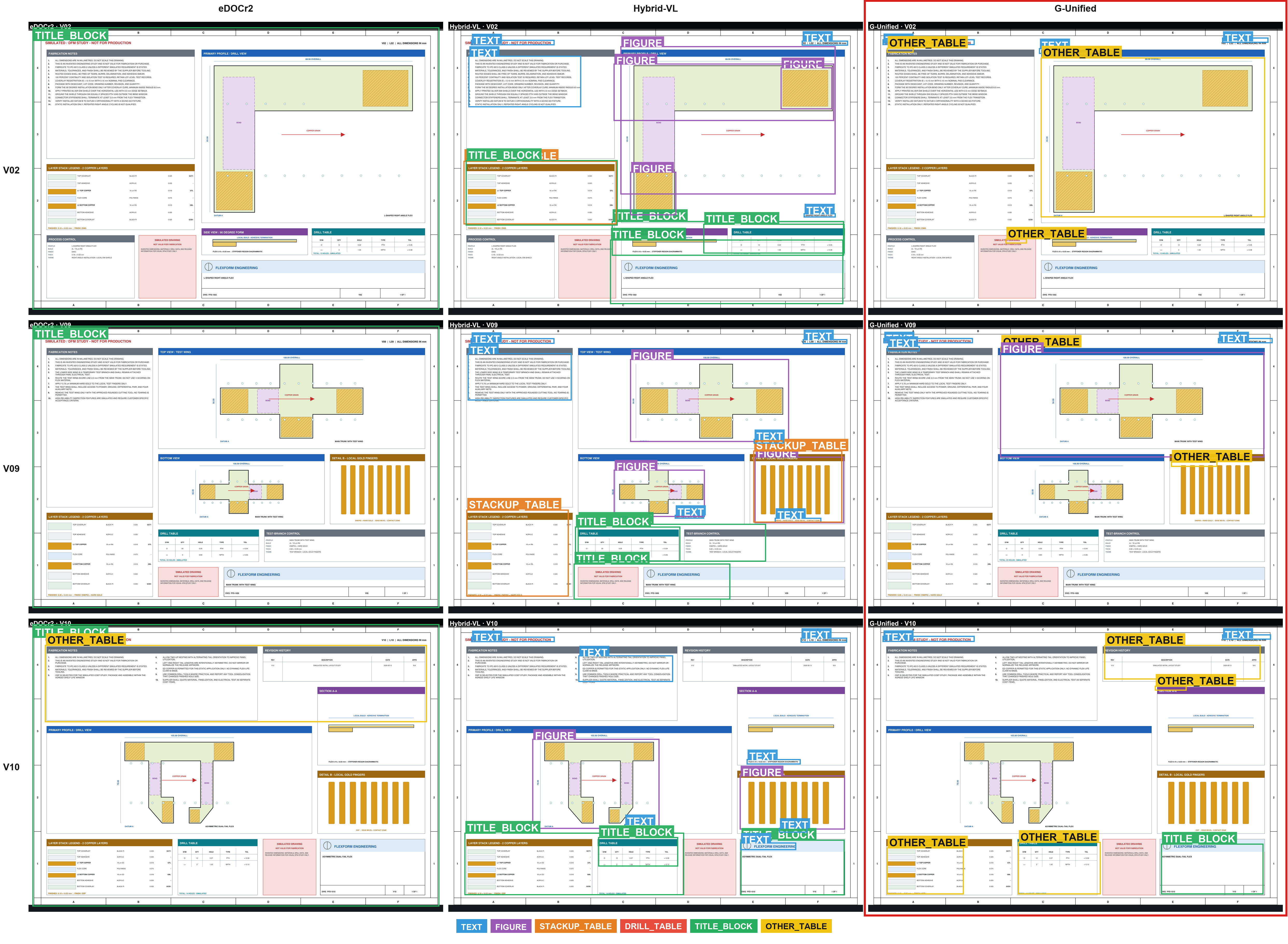}
  \caption{Qualitative localization comparison on three drawings from the ED dataset (V02, V09, and V10). Columns show eDOCr2, Hybrid-VL, and \modelname{}; colors follow the shared six-class ontology.}
  \label{fig:r2-qualitative}
\end{figure}

Figure~\ref{fig:r2-qualitative} compares eDOCr2, Hybrid-VL, and \modelname{} on three drawings from the ED dataset (V02, V09, and V10). The panels show differences in localization coverage and class assignment across page layouts. Aggregate claims use all 26 validation pages. Due to data privacy, all PCB models displayed here are generated by virtual engineering drawing.

\section{Conclusion}

This paper introduced G-Unified, a detector-free compact VLM that jointly predicts region classes, boxes, and content from full PCB drawings. On the fixed split of the ED dataset, Localization-First training improved strict localization F1 from 0.3942 to 0.4897. Although the reproduced Hybrid-VL localized better (F1 = 0.6704), G-Unified achieved the best content metrics among the unmodified end-to-end runs. Content recall remains the main bottleneck: only 27 of 78 content references were localized, yielding 23 usable outputs. eDOCr2 run yields usable content for 9, and parseable Florence outputs still give no exact match. Removing the detector--crop handoff simplifies the pipeline, but it does not remove the need to ground content correctly.These results are limited to one run per variant on 26 validation pages. Future work should improve content recall and structured decoding and evaluate broader PCB datasets.

\section*{Acknowledgments}

This work was partially supported by the Guangdong Provincial Key Laboratory of Interdisciplinary Research and Application for Data Science (Project Code 2022B1212010006); Beijing Normal--Hong Kong Baptist University (BNBU Research Grant Nos.~R0400001-22, UICR0400006-25, UICR0600048, and UICR0600036; Featured Innovation Project No.~2018KTSCX278); and Hong Kong aiKnow Limited. The authors also thank Junjie Liu of Trinity College Dublin for his valuable support.

\bibliographystyle{splncs04}
\bibliography{references}

\end{document}